\RequirePackage{fix-cm}
\documentclass[twocolumn]{svjour3}          
\smartqed  
\usepackage{graphicx}
\usepackage{latexsym}
\usepackage{breakcites}
\usepackage{amsmath,amssymb}
\usepackage{color}
\usepackage[font=scriptsize]{caption}
\usepackage{booktabs}
\usepackage{array}

\usepackage{savesym}
\savesymbol{vec}
\usepackage{newtxtext,newtxmath}
\restoresymbol{new}{vec}

\usepackage[caption = false]{subfig}
\usepackage{caption}
\usepackage[colorlinks=true]{hyperref}
\usepackage{multirow}
\DeclareMathOperator*{\argmin}{argmin} 

\newcolumntype{C}[1]{>{\centering\let\newline\\\arraybackslash\hspace{0pt}}m{#1}}
\journalname{IJCV special issue (Best papers of ECCV 2016)}
\begin{document}

\title{RED-Net: \\A Recurrent Encoder-Decoder Network for Video-based Face Alignment}

\titlerunning{RED-Net: A Recurrent Encoder-Decoder Network for Video-based Face Alignment}   

\author{Xi Peng         \and
        Rogerio S. Feris         \and
        Xiaoyu Wang         \and
        Dimitris N. Metaxas 
}
\institute{Xi Peng \at
              Rutgers University, Piscataway, NJ, 08854 \\
              Tel.: +1(917)803-7402\\
              \email{xpeng.cs@rutgers.edu}           
           \and
           Rogerio S. Feris \at
              IBM T. J. Watson Research Center, Yorktown Heights, NY, 10598 \\
              \email{rsferis@us.ibm.com}
           \and
           Xiaoyu Wang \at
           Intellifusion, Redmond, WA \\
           \email{fanghuaxue@gmail.com.com}
           \and
           Dimitris N. Metaxas \at
           Rutgers University, Piscataway, NJ, 08854 \\
           \email{dnm@cs.rutgers.edu}
}

\date{Submitted: April 19 2017 / Revised: December 12 2017}

\maketitle

\begin{abstract}
We propose a novel method for real-time face alignment in videos based on a recurrent encoder-decoder network model. Our proposed model predicts 2D facial point heat maps regularized by both detection and regression loss, while uniquely exploiting recurrent learning at both spatial and temporal dimensions. At the spatial level, we add a feedback loop connection between the combined output response map and the input, in order to enable iterative coarse-to-fine face alignment using a {\em single network model}, instead of relying on traditional cascaded model ensembles. At the temporal level, we first decouple the features in the bottleneck of the network into {\em temporal-variant factors}, such as pose and expression, and {\em temporal-invariant factors}, such as identity information. Temporal recurrent learning is then applied to the decoupled temporal-variant features. We show that such feature disentangling yields better generalization and significantly more accurate results at test time. We perform a comprehensive experimental analysis, showing the importance of each component of our proposed model, as well as superior results over the state of the art and several variations of our method in standard datasets. 

\keywords{Recurrent Learning, Encoder-Decoder Network, Face Alignment}
\end{abstract}

\section{Introduction}
Face landmark detection plays a fundamental role in many computer vision tasks, such as face recognition/verification, expression analysis, person identification, and 3D face modeling. It is also the basic technology component for a wide range of applications like video surveillance, emotion recognition, augmented reality on faces, etc. In the past few years, many methods have been proposed to address this problem, with significant progress being made towards systems that work in real-world conditions (``in the wild'').

Multiple lines of research have been explored for face alignment in last two decades. Early research includes methods based on active shape models (ASMs) \cite{Cootes92BMVC,StephenECCV08}  and active appearance models (AAMs) \cite{Gao2010}. ASMs iteratively deform a shape model to the target face image, while AAMs impose both shape and object appearance constraints in the optimization process. Recent advances in the field are largely driven by regression-based techniques \cite{XiongCVPR13,CaoIJCV14,ZhangECCV14,Lai2015,ZhangTangECCV14}. These methods usually take advantage of large-scale annotated training sets (lots of faces with labeled landmark points), achieving accurate results by learning discriminative regression functions that directly map facial appearance to landmark coordinates. The features extracted for regressing landmarks can be either hand-crafted features \cite{XiongCVPR13,CaoIJCV14}, or features extracted from convolutional neural networks \cite{ZhangECCV14,Lai2015,ZhangTangECCV14}. Although these methods can achieve very reliable results in standard benchmark datasets, they still suffer from limited performance in challenging scenarios, e.g., involving large face pose variations and heavy occlusions.

A promising direction to address these challenges is to consider video-based face alignment (i.e., sequential face landmark detection) \cite{ShenICCVW15,peng2017toward}, leveraging temporal information and identity consistency as additional constraints~\cite{WangCVPR16}. Despite the long history of research in rigid and non-rigid face tracking \cite{BlackCVPR95,OliverCVPR97,DecarloIJCV00,PatrasFG04}, current efforts have mostly focused on face alignment in still images \cite{SagonasICCVW13,ZhangECCV14,TzimiropoulosCVPR15,ZhuCVPR15}. When videos are considered as input, most methods perform landmark detection by independently applying models trained on still images in each frame in a tracking-by-detection manner \cite{WangTPAMI15}, with notable exceptions such as \cite{AsthanaCVPR14,PengICCV15,BMVC2016_129}, which explore incremental learning based on previous frames. These methods do not take full advantage of the temporal information to predict face landmarks for each frame. How to effectively model long-term temporal constraints while handling large face pose variations and occlusions is an open research problem for video-based face alignment.

In this work, we address this problem by proposing a novel recurrent encoder-decoder deep neural network model (see Figure \ref{fig:overview}), named as {\bf \emph{RED-Net}}. The encoding module projects image pixels into a low-dimensional feature space, whereas the decoding module maps features in this space to 2D facial point maps, which are further regularized by a regression loss.  

Our encoder-decoder framework allows us to explore spatial refining of our landmark prediction results, in order to handle faces with large pose variations. More specifically,  we introduce a feedback loop connection between the aggregated 2D facial point maps and the input. The intuition is similar to cascading multiple regression functions \cite{XiongCVPR13,ZhangECCV14} for iterative coarse-to-fine face alignment, but in our approach the iterations are modeled jointly with shared parameters, using a single network model. It provides significant parameter reduction when compared to traditional methods based on cascaded neural networks. A recurrent structure also avoids the effort to explicitly divide the task into multiple stage prediction problems.  This subtle difference makes the recurrent model more elegant in terms of holistic optimization. It can implicitly track the prediction behavior in different iterations for a specific face example, while cascaded predictions can only look at the immediate previous cascade stage. Our design also shares the same spirit of residual networks~\cite{he2016deep}. By adding feedback connections from the predicted heat map, the network only needs to implicitly predict the residual from previous predictions in subsequent iterations, which is arguably easier and more effective than directly predicting the absolute location of landmark points. 

For more effective temporal modeling, we first decouple the features in the bottleneck of the network into temporal-variant factors~\cite{peng2017reconstruction}, such as pose and expression, and temporal-invariant factors, such as identity. We disentangle the features into two components, where one component is used to learn face recognition using identity labels, and the other component encodes temporal-variant factors. To utilize temporal coherence in our framework, we apply recurrent temporal learning to the temporal-variant component. We used Long Short Term Memory (LSTM) to implicitly abstract  motion patterns by looking at multiple successive video frames, and use this information to improve landmark fitting accuracy. Landmarks with large pose variation are typically outliers in a landmark training set. By looking at multiple frames, it helps to reduce the inherent prediction variance in our model.

We show in our experiments that our encoder-decoder framework and its recurrent learning in both spatial and temporal dimensions significantly improve the performance of sequential face landmark detection. In summary, our work makes the following {\bf contributions}:

\begin{itemize}
\item We propose a novel recurrent encoder-decoder network model for real-time sequential face landmark detection. To the best of our knowledge, this is the first time a recurrent model is investigated to perform video-based facial landmark detection.
\item Our proposed {\em spatial recurrent learning} enables a novel iterative coarse-to-fine face alignment using a single network model. This is critical to handle large face pose changes and a more effective alternative than cascading multiple network models in terms of accuracy and memory footprint.
\item Different from traditional methods, we apply {\em temporal recurrent learning} to temporal-variant features which are decoupled from temporal-invariant features in the bottleneck of the network, achieving better generalization and more accurate results. 
\item We provide a detailed experimental analysis of each component of our model, as well as insights about key contributing factors to achieve superior performance over the state of the art. The project page is publicly available. \footnote{\scriptsize\url{https://sites.google.com/site/xipengcshomepage/eccv2016}}
\end{itemize}

\section{Related Work}
Face alignment has a long history of research in computer vision. Here we briefly discuss face alignment works related to our approach, as well as advances in deep learning, like the development of recurrent and encoder-decoder neural networks.

{\bf Regression-based face landmark detection.} Recently, regression-based face landmark detection methods~\cite{AsthanaCVPR13,SunCVPR13,XiongCVPR13,CaoIJCV14,ZhangECCV14,AsthanaCVPR14,ZhuCVPR15,TzimiropoulosCVPR15,JourablooCVPR16,WuCVPR16,ZhuCVPR16} have achieved significant boost in the generalization performance of face landmark detection, compared to algorithms based on statistical models such as Active shape models \cite{Cootes92BMVC,StephenECCV08} and Active appearance models~\cite{Gao2010}. 
Regression-based approaches directly regress landmark locations based on features extracted from face images. Landmark models for different points are learned either in an independent manner or in a joint fashion \cite{CaoIJCV14}. When all the landmark locations are learned jointly, implicit shape constraints are imposed because they share the same or partially the same regressors. 
This paper performs landmark detection via both a classification model and a regression model. Different from most previous methods, this work deals with face alignment in a video. It jointly optimizes detection output by utilizing multiple observations from the same person.

{\bf Cascaded models for landmark detection.} Additional accuracy improvement in face landmark detection performance can be obtained by learning cascaded regression models. Regression models from earlier cascade stages learn coarse detectors, while later cascade stages refine the result based on early predictions.  Cascaded regression helps to gradually reduce the prediction variance, thus making the learning task easier for later stage detectors. Many methods have effectively applied cascade-like regression models for the face alignment task~\cite{XiongCVPR13,SunCVPR13,ZhangECCV14}. The supervised descent method~\cite{XiongCVPR13} learns cascades of regression models based on SIFT features.
Sun \emph{et. al.}~\cite{SunCVPR13} proposed to use three levels of neural networks to predict landmark locations. 
Zhang \emph{et. al.}~\cite{ZhangECCV14} studied the problem via cascades of stacked auto-encoders which gradually refine the landmark position with higher resolution inputs. 
Compared to these efforts which explicitly define cascade structures, our method learns a spatial recurrent model which implicitly incorporates the cascade structure with shared parameters. It is also more "end-to-end" compared to previous works that divide the learning process into multiple stages. 


{\bf Face alignment in videos.} Most face alignment algorithms utilize temporal information by initializing the location of landmarks with detection results from the previous frame, performing alignment in a tracking-by-detection fashion ~\cite{WangTPAMI15}. Asthana \emph{et. al.}~\cite{AsthanaCVPR14} and Peng \emph{et. al.}~\cite{PengICCV15,BMVC2016_129} proposed to learn a person specific model using incremental learning.  However, incremental learning (or online learning) is a challenging problem, as the incremental scheme has to be carefully designed to prevent model drifting. In our framework, we do not update our model online. All the training is performed offline and we expect our LSTM unit to capture landmark motion correlations.

{\bf Recurrent neural networks.} Recurrent neural networks (RNNs) are widely employed in the literature of speech recognition~\cite{MikolovInterspeech10} and natural language processing~\cite{MikolovArxiv14}.
They have also been recently used in computer vision. For instance, in the tasks of image captioning~\cite{Karpathy_2015_CVPR} and video captioning~\cite{Yao_2015_ICCV}, RNNs are usually employed for text generation. RNNs are also popular as a tool for action classification. As an example, Veeriah \emph{et. al.}~\cite{VeeriahICCV15} use RNNs to learn complex time-series representations via high-order derivatives of states for action recognition. 

{\bf Encoder-decoder networks} Encoder and decoder networks are well studied in machine translation~\cite{ChoArxiv14} where the encoder learns the intermediate representation and the decoder generates the translation from the representation. It is also investigated in speech recognition~\cite{llu_is2015b} and computer vision~\cite{BadriCoRR15,HongCoRR15}. Yang \emph{et. al.}~\cite{YangNIPS15} proposed to decouple identity units and pose units in the bottleneck of the network for 3D view synthesis. However, how to fully utilize the decoupled units for correspondence regularization \cite{LongNIPS14} is still unexplored. In this work, we employ the encoder to learn a joint representation for identity, pose, expression as well as landmarks. The decoder translates the representation to landmark heatmaps. Our spatial recurrent model loops the whole encoder-decoder framework.

\section{Method}
\begin{figure*}[t]
\centering
\includegraphics[width=0.9\textwidth]{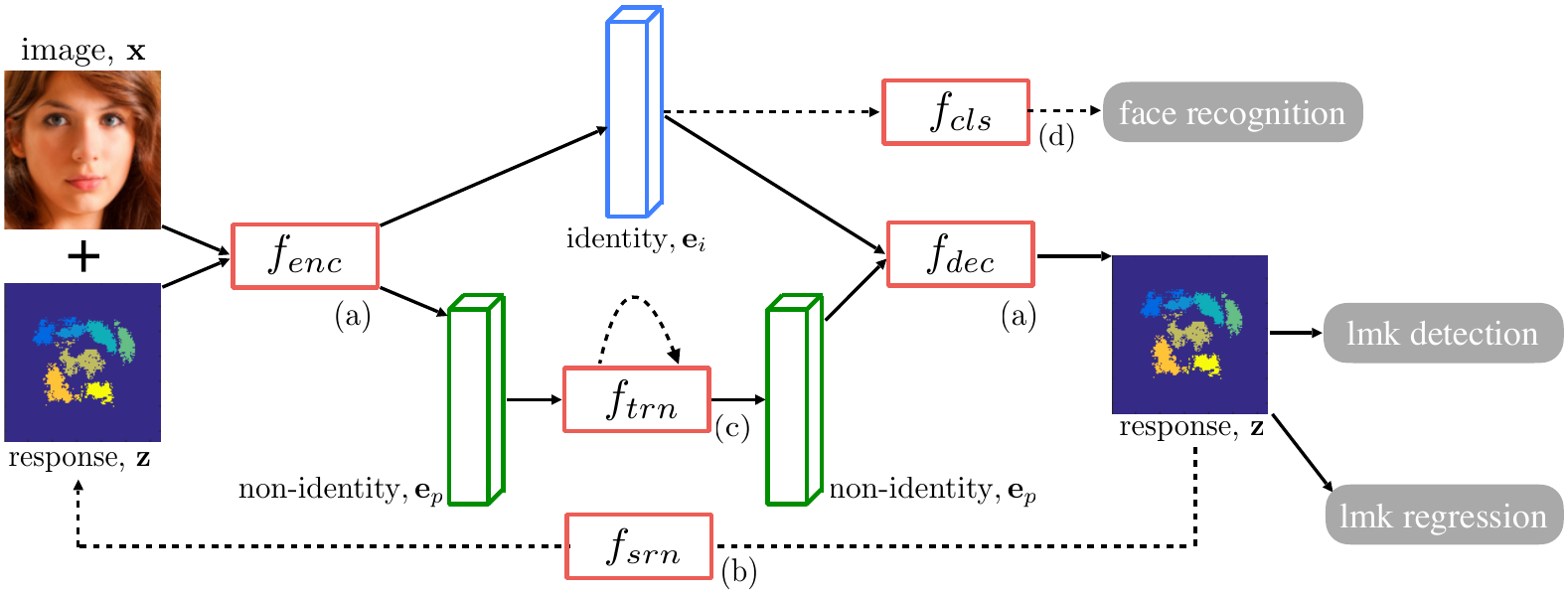}
\caption{Overview of the recurrent encoder-decoder network: {\bf (a)} encoder-decoder (Section \ref{sec:encdec}); {\bf (b)} spatial recurrent learning (Section \ref{sec:srn}); {\bf (c)} temporal recurrent learning (Section \ref{sec:trn}); and {\bf (d)} supervised identity disentangling (Section \ref{sec:cls}). $f_{enc},f_{dec},f_{srn},f_{trn},f_{cls}$ are potentially nonlinear and multi-layered mappings.} \label{fig:overview}
\end{figure*}

The task is to locate facial landmarks in sequential images using an end-to-end deep neural network. Figure \ref{fig:overview} shows an overview of our approach. The network consists of a series of nonlinear and multi-layered mappings, which can be functionally categorized as four modules: {\bf (1)} encoder-decoder $f_{enc}$ and $f_{dec}$, {\bf (2)} spatial recurrent learning $f_{srn}$, {\bf (3)} temporal recurrent learning $f_{trn}$, and {\bf (4)} constrained identity disentangling $f_{cls}$. Details of the novelty are described in following sections.

\subsection{Encoder-Decoder} \label{sec:encdec}

The input of the encoder-decoder is a single video frame $\mathbf{x} \in \mathbb{R}^{W \times H \times 3}$ and the output is a response map $\mathbf{z} \in \mathbb{R}^{W \times H \times C_z}$ which indicates landmark locations. $C_z = 7$ or $68$ depending on the number of landmarks to be predicted.

The \textit{encoder} performs a sequence of convolution, pooling and batch normalization \cite{IoffeCoRR15} to extract a low-dimensional representation $\mathbf{e}$ from both $\mathbf{x}$ and $\mathbf{z}$:
\begin{align}
 	\mathbf{e} = f_{enc}\big(\mathbf{x},\mathbf{z}; \theta_{enc}\big), \; f_{enc} : \mathbb{R}^{W \times H \times C} \rightarrow \mathbb{R}^{W_e \times H_e \times C_e},
\end{align}
where $f_{enc}\big(\cdot; \theta_{enc}\big)$ denotes the encoder mapping with parameters $\theta_{enc}$. We concatenate $\mathbf{x}$ and $\mathbf{z}$ along the channel dimension thus $C = 3 + C_z$. The concatenation is fed into the encoder as an updated input.

Symmetrically, the \textit{decoder} performs a sequence of unpooling, convolution and batch normalization to upsample the representation code to the response map:
\begin{align} \label{eq:z}
 	\mathbf{z} = f_{dec}(\mathbf{e}; \theta_{dec}), \; f_{dec} : \mathbb{R}^{W_e \times H_e \times C_e} \rightarrow \mathbb{R}^{W \times H \times C_z},
\end{align}
where $f_{dec}\big(\cdot; \theta_{dec}\big)$ denotes the decoder mapping with parameters $\theta_{dec}$. {$\mathbf{z}$ has the same $W \times H$ dimension as $\mathbf{x}$ but $C_z$ channels for $C_z$ landmarks. Each channel presents pixel-wise confidences of the corresponding landmark.

The encoder-decoder design plays an important role in our task. {\bf First}, the decoder's output $\mathbf{z}$ has the same resolution (but a different number of channels) as the input image $\mathbf{x}$. Thus it is easy to directly concatenate $\mathbf{z}$ with $\mathbf{x}$ along the channel dimension. The concatenation provides pixel-wise spatial cues to update the landmark prediction by the proposed {\em spatial recurrent learning} ($f_{srn}$). We will explain it soon in Section \ref{sec:srn}.

{\bf Second}, the encoder-decoder network can achieve a low-dimensional representation $\mathbf{e}$ in the bottleneck. We can utilize the domain prior to decouple $\mathbf{e}$ into two parts: the identity code $\mathbf{e}_{i}$, which is temporal-invariant as we are tracking the same person; and the non-identity code $\mathbf{e}_{p}$, which models temporal-variant factors such as head pose, expression, illumination, and etc.

In Section \ref{sec:trn}, we propose the {\em temporal recurrent learning} ($f_{trn}$) to model the changes of $\mathbf{e}_{p}$. In Section \ref{sec:cls}, we show how to speed up the network training by carrying out the {\em supervised identity disentangling} ($f_{cls}$) on $\mathbf{e}_{i}$.

{\bf Third},the encoder-decoder network enables a fully convolutional design. The bottleneck embedding $\mathbf{e}$ and output response map $\mathbf{z}$ are feature maps instead of fully-connected neurons that are often used in ordinary convolutional neural networks. This design is highly memory-efficient and can significantly speed up the training and testing \cite{LongCoRR14}, which is preferred by video-based applications.

\begin{figure*}[t]
\minipage{0.49\textwidth}
  \includegraphics[width=0.95\linewidth]{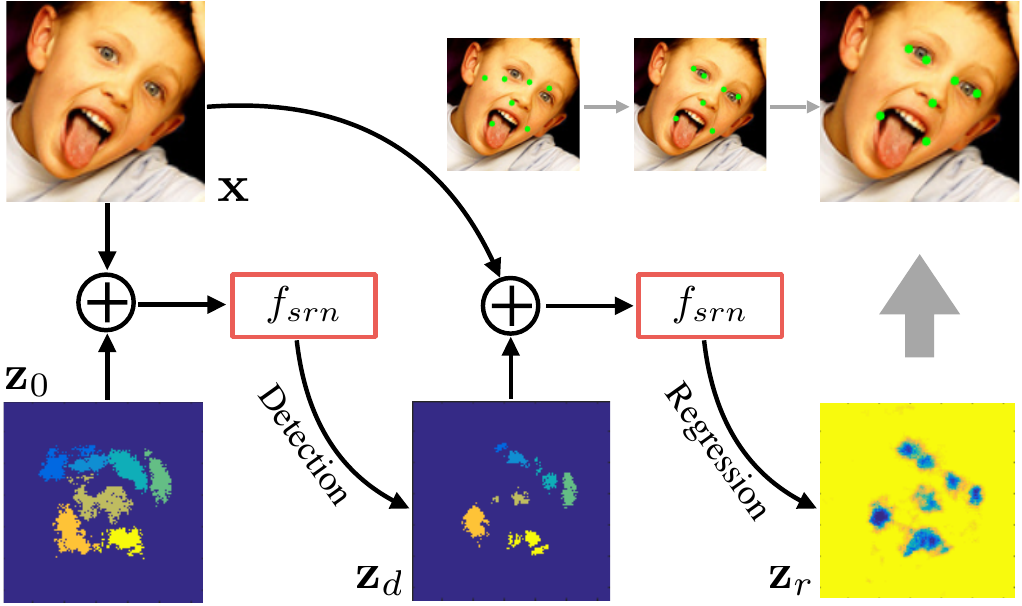}
  \centering
  \caption{An unrolled illustration of {\em spatial recurrent learning}. The response map is pretty coarse when the initial guess is far away from the ground truth if large pose and expression exist. It eventually gets refined in the successive recurrent steps.
  }\label{fig:fig_fsrn}
\endminipage\hfill
\minipage{0.49\textwidth}
  \includegraphics[width=0.9\linewidth]{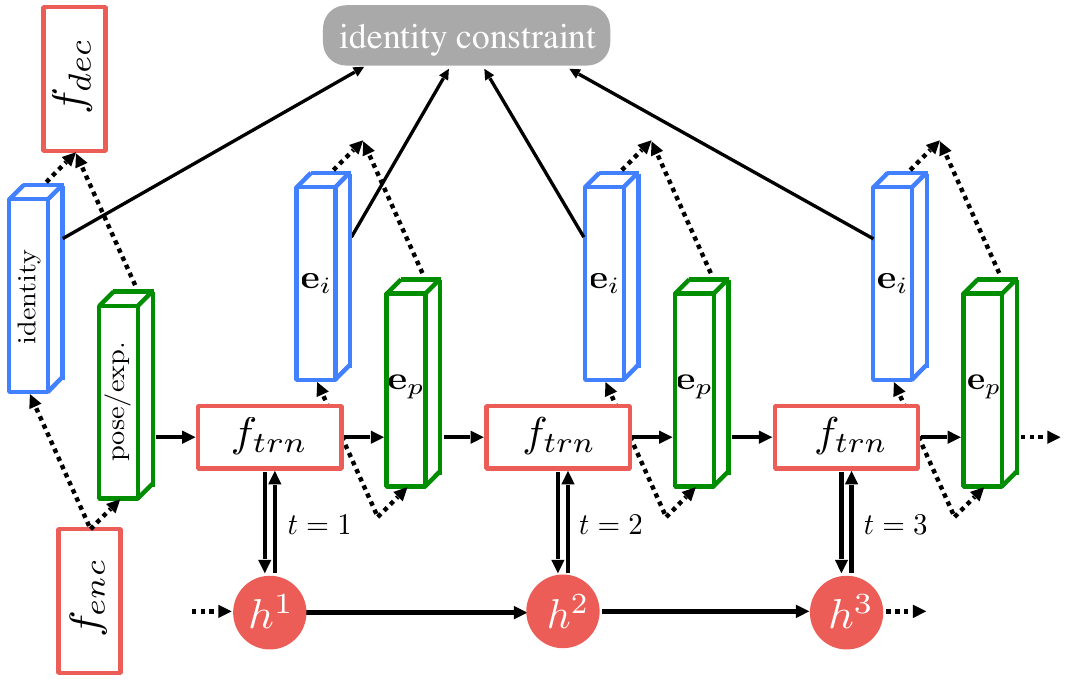}
  \centering
  \caption{An unrolled illustration of {\em temporal recurrent learning}. $\mathcal{C}_{i}$ encodes temporal-invariant factor which subjects to the same identity constraint. $\mathcal{C}_{p}$ encodes temporal-variant factors which is further modeled in $f_{trn}$.}\label{fig:fig_ftrn}
\endminipage\hfill
\end{figure*}

\subsection{Spatial Recurrent Learning} \label{sec:srn}

The purpose of spatial recurrent learning is to pinpoint landmark locations in a coarse-to-fine manner. Unlike existing approaches \cite{SunCVPR13,ZhangECCV14} that employ multiple networks in cascade, we accomplish the coarse-to-fine search in a single network in which the parameters are jointly learned in successive recurrent steps.

The spatial recurrent learning is performed by iteratively feeding back the previous prediction, stacked with the image as shown in Figure~\ref{fig:fig_fsrn}, to eventually push the shape prediction from an initial guess to the ground truth:
\begin{align} \label{eq:srn}
	\mathbf{z}_{k} = f_{srn}\big(\mathbf{x},\mathbf{z}_{k-1}; \theta_{srn}\big), \; k=1,\cdots,K
\end{align}
where $f_{srn}\big(\cdot; \theta_{srn}\big)$ denotes the spatial recurrent mapping with parameters $\theta_{srn}$. $\mathbf{z}_0$ is the initial response map, which could be a response map generated by the mean shape or the output of the previous frame.

In our conference version \cite{peng2016recurrent}, detection-based supervision is performed in every recurrent step. It is robust to appearance variations but lacks precision, because pixels within a certain radius around the ground-truth location are labeled using the same value. To address this limitation, motivated by \cite{Bulat2016}, we propose to further explore the spatial recurrent learning by performing detection-followed-by-regression in successive steps.

Specially, we carry out a two-step recurrent learning by setting $K=2$. The first step performs {\em landmark detection} that aims to locate 7 major facial components ({\em i.e.} $C=7$ in Equation \eqref{eq:z}). The second step performs {\it landmark regression} that refines all 68 landmarks positions ({\em i.e.} $C=68$). For clarity, we use $C_d$ and $C_r$ to denote the number of channels output by the detection and the regression steps, respectively. 

The landmark detection step guarantees fitting robustness especially in large pose and partial occlusions. The encoder-decoder aims to output a binary map of $C_{d}$ channels, one for each major facial component. The detection step outputs:
\begin{align}
	\mathbf{z}_{d} = f_{dec} \big( f_{enc}(\mathbf{x},\mathbf{z}_0; \theta_{enc}); \theta_{dec} \big), \; \mathbf{z}_{d} \in \mathbb{R}^{W \times H \times C_{d}},
\end{align}
where the detection task can be trained using pixel-wise sigmoid cross-entropy loss function:
\begin{align} \label{eq:det_loss}
	\ell_{d} = \frac{1}{M_d} \sum_{c=1}^{C_d} \sum_{i=1}^{W} \sum_{j=1}^{H} z_{ij}^c log y_{ij}^c + (1 - z_{ij}^c) log (1 - y_{ij}^c),
\end{align}
where $M_d = C_d \times W \times H$. Here $z_{ij}^c$ denotes the sigmoid output at pixel location $(i,j)$ in $\mathbf{z}_{d}$ for the $c$-th landmark. $y_{ij}^c$ is the ground-truth label at the same location, which is set to 1 to mark the presence of the corresponding landmark and 0 for the remaining background.

Note that this loss function is different from the N-way cross-entropy loss used in our previous conference paper \cite{peng2016recurrent}. It allows multiple class labels for a single pixel, which helps to tackle the landmark overlaps.  

The landmark regression step improves the fitting accuracy from the outputs of the previous detection step. The encoder-decoder aims to output a heatmap of $C_{r}$ channels, one for each landmark. The regression step outputs:
\begin{align}
	\mathbf{z}_{r} = f_{dec} \big( f_{enc}(\mathbf{x},\mathbf{z}_{det}; \theta_{enc}); \theta_{dec} \big), \; \mathbf{z}_{r} \in \mathbb{R}^{W \times H \times C_{r}},
\end{align}
where the regression task can be trained using pixel-wise $L_2$ loss function:
\begin{align} \label{eq:reg_loss}
	\ell_{r} = \frac{1}{M_r} \sum_{c=1}^{C_r} \sum_{i=1}^{W} \sum_{j=1}^{H} \| z_{ij}^c - y_{ij}^c \|^2_2,
\end{align}
where $M_r = C_d \times W \times H$. Here $z_{ij}^c$ denotes the heatmap value of the $c$-th landmark at pixel location $(i,j)$ in $\mathbf{z}_{r}$ for the $c$-th landmark. $y_{ij}^c$ is the ground-truth value at the same location, which obeys a Gaussian distribution centered at the landmark with a pre-defined standard deviation.

Now the spatial recurrent learning (Equation \eqref{eq:srn}) can be achieved by minimizing the detection loss (Equation \eqref{eq:det_loss}) and the regression loss (Equation \eqref{eq:reg_loss}), simultaneously:
\begin{align} \label{eq:srnloss}
	\argmin_{\theta_{enc},\theta_{dec}} \ell_{d} + \lambda \ell_{r},
\end{align}
where $\lambda$ balances the loss between the two tasks. Note that the spatial recurrent learning do not introduce new parameters but sharing the same parameters of the encoder-decoder network, {\em i.e.} $\theta_{srn} = \{\theta_{enc},\theta_{dec}\}$.

The spatial recurrent learning is highly memory efficient. It is capable of end-to-end training, which is a significant advantage compared with the cascade framework \cite{Bulat2016}. More importantly, the network can jointly learn the coarse-to-fine fitting strategy in recurrent steps, instead of training cascaded networks independently \cite{SunCVPR13,ZhangECCV14}, which guarantees robustness and accuracy in challenging conditions.

\begin{table*}[h]
\centering
\caption{Specification of the VGGNet-based $f_{enc/dec}$ design: block name ({\bf Top}), feature map dimension ({\bf Middle}), and layer configuration ({\bf Bottom}). $\big[3\times3,64\big]$ means there are 64 filters (channels), each has a size of $3\times3$. Pooling or unpooling operations are performed after or before each module. The pooling window is $2\times2$ with a stride of $2$.
} \label{tab:encdec_vgg}
\begin{tabular}{c | c | c | c | c | c | c | c | c}
\toprule
$A_0$ & $A_1$ & $A_2$ & $A_3$ & $A_4$ & $B_4$ & $B_3$ & $B_2$ & $B_1$ \\
\hline
$128\times128$ & $64\times64$  & $32\times32$ & $16\times16$ & $8\times8$ & $16\times16$  & $32\times32$ & $64\times64$ & $128\times128$ \\
\hline
$2 \times$ conv & $2 \times$ conv & $3 \times$ conv & $3 \times$ conv & $3 \times$ conv & unpooling & unpooling & unpooling & unpooling \\
$\big[3\times3,64\big]$ & 
$\big[3\times3,128\big]$ & 
$\big[3\times3,256\big]$ & 
$\big[3\times3,512\big]$ & 
$\big[3\times3,512\big]$ & 
$3 \times$ conv & $3 \times$ conv & $3 \times$ conv & $2 \times$ conv \\
pooling & pooling & pooling & pooling & - & 
$\big[3\times3,512\big]$ & 
$\big[3\times3,512\big]$ & 
$\big[3\times3,256\big]$ & 
$\big[3\times3,128\big]$ \\
\bottomrule
\end{tabular}
\end{table*}

\begin{figure*}[t]
\centering
\includegraphics[width=0.85\textwidth]{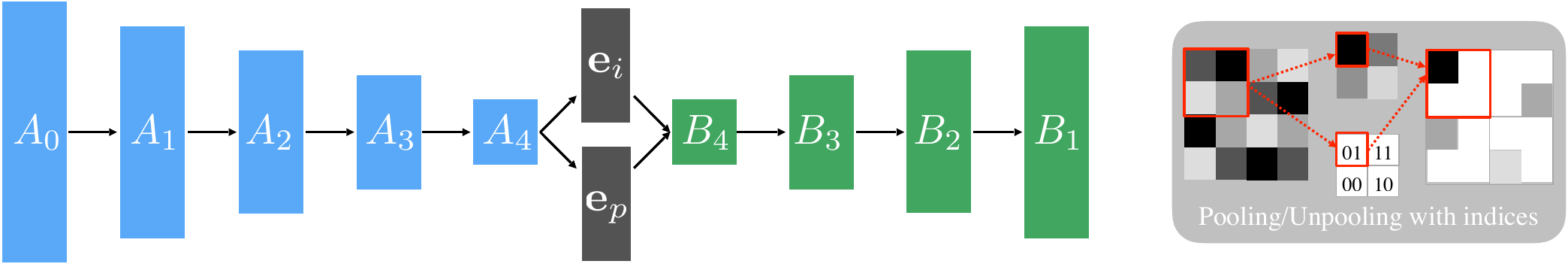}
\caption{
{\bf Left:} the architecture of the VGGNet-based $f_{enc/dec}$ design. The encoder ($A_{0-4}$) and the decoder ($B_{4-1}$) are nearly symmetrical except that $f_{enc}$ has one more block $A_0$. $A_0$ downsamples the input image from $256\times256$ to $128\times128$. So $\mathbf{x}$ and $\mathbf{z}$ have the same resolution and can be easily concatenated along the channel dimension. {\bf Right:} an illustration of the pooling/unpooing with indices. The corresponding pooling and unpooling share pooling indices using a 2-bit switch in each 2 $\times$ 2 pooling window.
}\label{fig:encdec_vgg}
\end{figure*}

\subsection{Temporal Recurrent Learning} \label{sec:trn}

In addition to the spatial recurrent learning, we also propose a temporal recurrent learning to model factors, {\em e.g.} head pose, expression, and illumination, that may change over time. These factors affect the landmark locations significantly \cite{PengCVIU15}. Thus we can expect improved tracking accuracy by modeling their temporal variations.

As mentioned in Section \ref{sec:encdec}, the bottleneck embedding $\mathbf{e}$ can be decoupled into two parts: the identity code $\mathbf{e}_i$ and the non-identity code $\mathbf{e}_p$:
\begin{align}
 	\mathbf{e}_{i} \in \mathbb{R}^{W_e \times H_e \times C_i}, \mathbf{e}_{p} \in \mathbb{R}^{W_e \times H_e \times C_p}, C_e = C_i + C_p,  
\end{align}
where $\mathbf{e}_i$ and $\mathbf{e}_p$ model the temporal-invariant and -variant factors, respectively. We leave $\mathbf{e}_i$ to Section \ref{sec:cls} for additional identity supervision, and exploit variations of $\mathbf{e}_{p}$ via the recurrent model. Please refer to Figure \ref{fig:fig_ftrn} for an unrolled illustration of the proposed temporal recurrent learning.

Mathematically, given $T$ successive video frames $\{\mathbf{x}^{t}; t=1, \cdots, T\}$, the encoder extracts a sequence of embeddings $\{\mathbf{e}^{t}_i, \mathbf{e}^{t}_p; t=1, \cdots, T\}$. Our goal is to achieve a nonlinear mapping $f_{trn}$, which simultaneously tracks a latent state $h^t$ and updates $\mathbf{e}_{p}^t$ at time $t$:
\begin{align}
	h^t & = p(\mathbf{e}_{p}^t, h^{t-1}; \theta_{trn}),  \;\;\; t=1,\cdots,T \nonumber \\
	\mathbf{e}_{p}^{t*} & = q(h^t; \theta_{trn}), 
\end{align}
where $p(\cdot)$ and $q(\cdot)$ are functions of $f_{trn}\big(\cdot; \theta_{trn}\big)$ with parameters $\theta_{trn}$. $\mathbf{e}_{p}^{t*}$ is the update of $\mathbf{e}_{p}^t$.

The temporal recurrent learning is trained using $T$ successive frames. At each frame, the detection and regression tasks are performed for the spatial recurrent learning. The recurrent learning is performed by minimizing Equation \eqref{eq:srnloss} at every time step $t$:
\begin{align} \label{eq:trnloss}
	\argmin_{\theta_{enc},\theta_{dec},\theta_{trn}} \sum_{t=1}^{T} \ell_{d}^t + \lambda \ell_{r}^t,
\end{align}
where $\theta_{trn}$ denotes network parameters of the temporal recurrent learning, {\em e.g.} parameters of LSTM units. It is worth mentioning that, we perform recurrent learning in both spatial and temporal dimensions by jointly optimizing $\{\theta_{enc}, \theta_{dec}, \theta_{trn}\}$ in Equation \eqref{eq:trnloss}.

The temporal recurrent module is memorizing as well as modeling the changing pattern of the temporal-variant factors. Our experiments indicated that the offline learned model can significantly improve the online fitting accuracy and robustness, especially when large variations or partial occlusions happen.

\subsection{Supervised Identity Disentangling} \label{sec:cls}
There is no guarantee that temporal-invariant and -variant factors can be completely decoupled in the bottleneck by simply splitting the bottleneck representation $\mathbf{e}$ into two parts \cite{peng2017reconstruction}. More supervised information is required to achieve the disentangling. To address this issue, we propose to apply a face recognition task on the identity code $\mathbf{e}_{i}$, in addition to the temporal recurrent learning applied on non-identity code $\mathbf{e}_{p}$.

The supervised identity disentangling is formulated as an $N$-way classification problem. $N$ is the number of unique individuals present in the training sequences. In general, we associate the identity representation $\mathbf{e}_{i}$ with a one-hot encoding $\mathbf{z}_{i}$ to indicate the score of each identity:
\begin{align}
 	\mathbf{z}_{i} = f_{cls}(\mathbf{e}_{i}; \theta_{cls}), \; f_{cls} : \mathbb{R}^{W_e \times H_e \times C_i} \rightarrow \mathbb{R}^{N},
\end{align}
where $f_{cls}(\cdot; \theta_{cls})$ is the identity classification mapping with parameters $\theta_{cls}$. The identity task is trained using $N$-way cross-entropy loss:
\begin{align} \label{eq:clsloss}
	\ell_{cls} = \frac{1}{N} \sum_{n=1}^{N} z^n log y^n + (1 - z^n) log (1 - y^n),
\end{align}
where $z^n$ denotes the softmax activation of the $n$-th element in $\mathbf{z}_{i}$. $y^n$ is the $n$-th element of the identity annotation $\mathbf{y}_{i}$, which is a one-hot vector with a $1$ for the correct identity and all $0$s for others.

Now we can jointly train all the three tasks, {\em i.e.} $f_{srn}$, $f_{trn}$, and $f_{cls}$. Based on Equation \eqref{eq:trnloss} and \eqref{eq:clsloss}, we simultaneously minimize the detection and regression loss together with the identity loss at every time step $t$:
\begin{align} \label{eq:allloss}
	\argmin_{\theta_{enc},\theta_{dec},\theta_{trn},\theta_{cls}} \sum_{t=1}^{T} \ell_{det}^t + \lambda \ell_{reg}^t + \gamma \ell_{cls}^t,
\end{align}
where $\gamma$ weights the identity constraint. An obvious advantage of our approach is that the whole network can be trained end-to-end by optimizing all parameters $\{\theta_{enc},\theta_{dec},\theta_{trn},\theta_{cls}\}$ simultaneously, which guarantees an efficient learning. 

It has been shown in \cite{ZhangTangECCV14} that learning the face alignment task together with correlated tasks, \textit{e.g.} head pose, can improve the fitting performance. We have a similar observation when adding face recognition task to the alignment task. More importantly, we find that the additional identity task can effectively speed up the training of the entire encoder-decoder network. In addition to more supervision, the identity task helps to decouple the identity and non-identity factors more completely, which facilitates the training of the temporal recurrent learning.

\begin{table*}[t]
\centering
\caption{Specification of the ResNet-based $f_{enc/dec}$ design: block name ({\bf Top}), feature map dimension ({\bf Middle}), and layer configuration ({\bf Bottom}). We use conv/decov layers with a stride of $2$ to halve or double the feature map dimensions. Thus no pooling/unpooling layer is used. The skip connections $E_{1-3}$ are specified in Table \ref{tab:encdec_skip}.} \label{tab:encdec_res}
\begin{tabular}{c | c | c | c | c | c | c | c | c}
\toprule
$C_0$ & $C_1$ & $C_2$ & $C_3$ & $C_4$ & $D_4$ & $D_3$ & $D_2$ & $D_1$ \\
\hline
$128\times128$ & $64\times64$  & $32\times32$ & $16\times16$ & $8\times8$ & $16\times16$  & $32\times32$ & $64\times64$ & $128\times128$ \\
\hline
$1\times$ conv & $3\times$ conv & $8\times$ conv & $36\times$ conv & $3\times$ conv & $1\times$ dconv & $1\times$ dconv & $1\times$ dconv & $1\times$ dconv \\
$\begin{bmatrix} 7\times7,64 \\ \text{strid},2 \end{bmatrix}$ & 
$\begin{bmatrix} 1\times1,64 \\ 3\times3,64 \\ 1\times1,256 \end{bmatrix}$ & 
$\begin{bmatrix} 1\times1,128 \\ 3\times3,128 \\ 1\times1,512 \end{bmatrix}$ & 
$\begin{bmatrix} 1\times1,256 \\ 3\times3,256 \\ 1\times1,1024 \end{bmatrix}$ & 
$\begin{bmatrix} 1\times1,512 \\ 3\times3,512 \\ 1\times1,2048 \end{bmatrix}$ & 
$\begin{bmatrix} 2\times2,512 \\ \text{stride},2 \\ 1\times1,1024 \end{bmatrix}$ & 
$\begin{bmatrix} 2\times2,256 \\ \text{stride},2 \\ 1\times1,512 \end{bmatrix}$ & 
$\begin{bmatrix} 2\times2,128 \\ \text{stride},2 \\ 1\times1,256 \end{bmatrix}$ & 
$\begin{bmatrix} 2\times2,64 \\ \text{stride},2 \\ 1\times1,128 \end{bmatrix}$ \\
\bottomrule
\end{tabular}
\end{table*}

\begin{figure*}[th]
\centering
\includegraphics[width=0.8\textwidth]{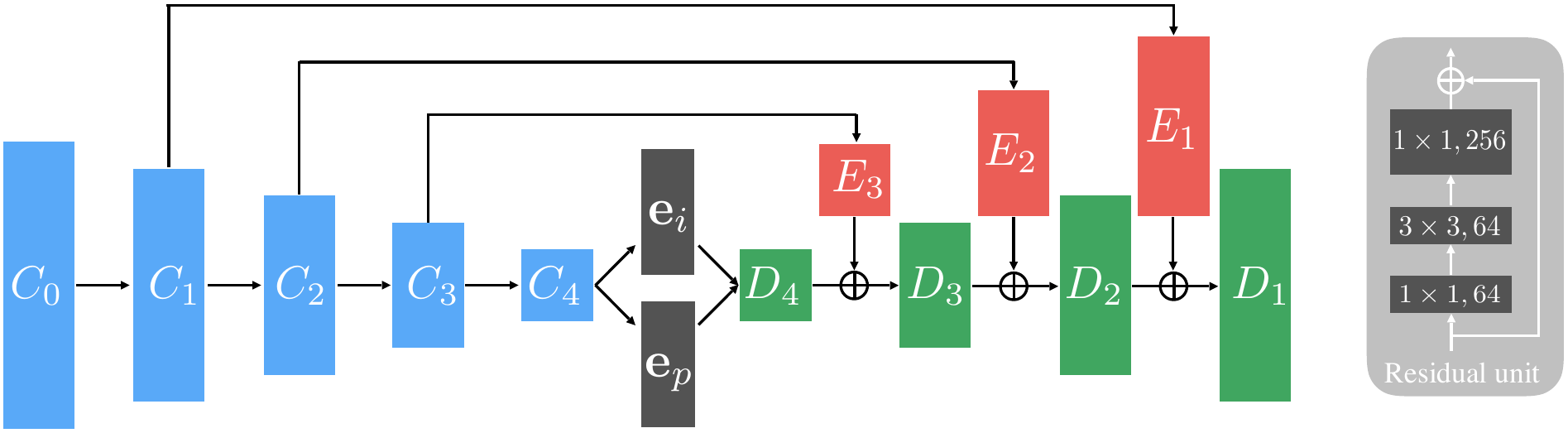}
\caption{{\bf Left:} the architecture of ResNet-based $f_{enc/dec}$ design ({\bf Left}). The encoder ($C_{0-4}$) and the decoder ($D_{4-1}$) are asymmetrical. $f_{enc}$ is much deeper than $f_{dec}$, {\it i.e.} 151 vs. 4 layers. $C_0$ downsamples the input image from $256\times256$ to $128\times128$. Skip connections ($E_{1-3}$) are used to bridge hierarchical spatial information at different resolutions. {\bf Right:} an example of residual unit used in $C_1$. $1\times1$ convolutional layers are used in the residual unit to cut down the number of filter parameters.}\label{fig:encdec_res}
\end{figure*}

\section{Network Architecture} \label{sec:arch}
We present the architecture details of proposed modules: $f_{enc/dec}$, $f_{srn}$, $f_{trn}$, and $f_{cls}$. All the four modules are designed in a single network that can be trained end-to-end. We first introduce two variant designs of $f_{enc/dec}$, based on which $f_{srn}$, $f_{trn}$, and $f_{cls}$ are designed accordingly.

\subsection{The Design of $f_{enc}$ and $f_{dec}$} \label{sec:arch_encdec}
To best evaluate the proposed method, we investigate two variant designs of the encoder-decoder: VGGNet \cite{SimonyanCoRR14} based and ResNet \cite{he2016deep} based. The VGGNet-based design has a symmetrical structure between the encoder and decoder; while the ResNet-based design has an asymmetrical structure due to the usage of the residual modules.

{\bf VGGNet-based design.} Table \ref{tab:encdec_vgg} presents the network specification. Figure \ref{fig:encdec_vgg} (left) shows the network architecture. The encoder is designed based on a variant of the VGG-16 network \cite{SimonyanCoRR14,KendallCoRR15}. It has 13 convolutional layers of constant $3 \times 3$ filters. We can, therefore, initialize the training process from weights trained on large datasets for object classification. We remove all fully connected layers in favor of a fully convolutional manner \cite{LongCoRR14}, which can effectively reduce the number of parameters from 117M to 14.8M \cite{BadriCoRR15}. The bottleneck feature maps are split into two parts for the identity and non-identity codes, respectively. This design preserves rich spatial information in 3D feature maps rather than 1D feature vectors, which is important for landmark localization.

We use max-pooling to halve the feature resolution at the end of each convolutional block. The pooling window size is $2\times2$ and the stride is $2$. Although max-pooling can help to achieve translation invariance, it would cause a considerable loss of spatial information especially when multiple max-pooling layers are applied in a cascade. To solve this issue, we use a 2-bit code to record the index of the maximum activation selected in a $2\times2$ pooling window \cite{ZeilerECCV14}. As illustrated in Figure \ref{fig:encdec_vgg} (right), the memorized index is then used in the corresponding unpooling layer to place each activation back to its original location. This strategy is particularly useful when the decoder recovers the input structure from highly compressed feature maps. Besides, it is more efficient to store spatial indices than to memorize entire feature maps of float precision, which is a common setup in FCNs \cite{LongCoRR14}.

The decoder is nearly symmetrical to the encoder with a mirrored configuration but replacing all max-pooling with unpooling layers. The encoder is slightly deeper than the decoder with one more covolutional block $A_0$ at the beginning. $A_0$ downsamples the input image from $256\times256$ to $128\times128$. So $\mathbf{x}$ and $\mathbf{z}$ have the same resolution and can be easily concatenated along the channel dimension. We find that batch normalization \cite{IoffeCoRR15} can significantly boost the training speed since it reduces internal shifting in the mini batch. Thus, we apply batch normalization as well as rectified linear unit (ReLU) \cite{NairICML10} after each convolutional layer.


{\bf ResNet-based design.} Table \ref{tab:encdec_res} presents the network specification. Figure \ref{fig:encdec_res} (left) shows the network architecture. The encoder is designed based on a variant of the ResNet-152 \cite{he2016deep}, which has 50 residual units of totally 151 convolutional layers. Figure \ref{fig:encdec_res} (right) shows a residual unit used in $C_1$. $1\times1$ convolutional layers are used to cut down the number of filter parameters. According to \cite{he2016deep}, the residual shortcut guarantees efficient training of the very deep network, as well as improved performance compared with vanilla design \cite{SimonyanCoRR14}. Stride-2 convolutions instead of max poolings are used to halve the feature map resolution at the end of each block.

Different from the VGGNet-based design, the encoder and decoder are asymmetrical. The encoder is much deeper than the decoder, and the decoder has only 4 upsampling blocks of totally 4 convolutional layers. A practical consideration behind this design is that the encoder has to tackle a complicated task, {\it e.g.} understand the image and translate it to a low-dimensional embedding, while the decoder's task is relatively simpler, {\it e.g.} recover a set of response maps to mark landmark locations from the embedding. We use stride-2 de-convolutions to double the feature map resolution in each block. Similar to the VGGNet-based design, an additional convolutional block $C_0$ is used to downsample the inpuy image from $256\times256$ to $128\times128$. So $\mathbf{x}$ and $\mathbf{z}$ have the same resolution for an easy channel-wise concatenation.

Another difference between the ResNet-based design and the VGGNet-based design is the usage of skip connections $E_{3-1}$ \cite{OhNIPS15} as shown in Figure \ref{fig:encdec_res} and specified in Table \ref{tab:encdec_skip}. These skip connections are used to bridge hierarchical spatial information between the encoder and decoder at different resolutions. They provide shortcuts of the gradient flow in backpropagation for efficient training. Besides, they also enable us to use a shallow decoder design since rich spatial information can be delivered through the shortcuts.

\begin{table}[t]
\centering
\caption{Specification of the skip connections. Note that $E_3$ and $C_1$, $E_2$ and $C_2$, $E_1$ and $C_1$ share the same configurations. The bridged features are directly added to the outputs of $D_{4-1}$ at the corresponding resolutions.} \label{tab:encdec_skip}
\begin{tabular}{c | c | c }
\toprule
$E_3$ & $E_2$ & $E_1$  \\
\hline
$16\times16$ & $32\times32$  & $64\times64$ \\
\hline
$3\times$ conv & $3\times$ conv & $3\times$ conv \\
$\begin{bmatrix} 1\times1,256 \\ 3\times3,256 \\ 1\times1,1024 \end{bmatrix}$ & 
$\begin{bmatrix} 1\times1,128 \\ 3\times3,128 \\ 1\times1,512 \end{bmatrix}$ & 
$\begin{bmatrix} 1\times1,64 \\ 3\times3,64 \\ 1\times1,256 \end{bmatrix}$ \\
\bottomrule
\end{tabular}
\end{table}

\subsection{The Design of $f_{srn}$ and $f_{trn}$}
The design of the proposed $f_{srn}$ and $f_{trn}$ aims to tradeoff between network complexity and training or testing efficiency.

{\bf Spatial recurrent learning.} We perform a two-step spatial recurrent learning. Particularly, the first step performs landmark detection to locate 7 major facial components that are robust to variations, {\em i.e.} four corners of left/right eyes, one nose tip, and two corners of the mouth. The second step performs landmark regression to refine the predicted locations of all 68 landmarks. This coarse-to-fine strategy guarantees efficient and robust spatial recurrent learning.

As mentioned in Section \ref{sec:srn}, the landmark detection task outputs a binary map of $C_d=7$ channels, in which the values within a radius of 5 pixels around the ground truth are set to 1 and the values for the remaining background are set to 0. The landmark regression task outputs a heat map of $C_r=68$ channels, in which the correct locations are represented by Gaussian with a standard deviation of 5 pixels. The two tasks share the weights of the entire encoder-decoder except for the last convolutional layer, which uses $1\times1$ convolutional layers to adapt to either the binary map or the heat map.

In either landmark detection or regression, the foreground pixels are much less than the background ones, which lead to highly unbalanced loss contributions. To solve this issue, we enlarge the foreground loss defined in Equation \eqref{eq:srnloss} and \eqref{eq:trnloss} by multiplying a constant weight (15 in most cases) to focus more on foreground pixels.

{\bf Temporal recurrent learning.} We specify the configuration of $f_{trn}$ in Figure \ref{fig:trn_cls} (left). A Long Short Term Memory (LSTM) module \cite{HochreiterNC97,OhNIPS15} is used to model the temporal variations of the identity code. There are $256$ hidden neurons are used in LSTM. We empirically set the number of successive frames as $T=10$ in Equation \eqref{eq:trnloss}. The prediction loss is calculated at each time step. Directly feeding the non-identity code $\mathbf{e}_{p}$ into LSTM layers would lead to a slow training as it needs a large number of neurons for both the input and output. Instead, we apply average pooling to compress $\mathbf{e}_{p}$ to a $256d$ vector before inputting to the LSTM and recover it by unpooling with indices as shown in Figure \ref{fig:encdec_vgg} (left).


\begin{figure}[t]
\centering
\includegraphics[width=0.48\textwidth]{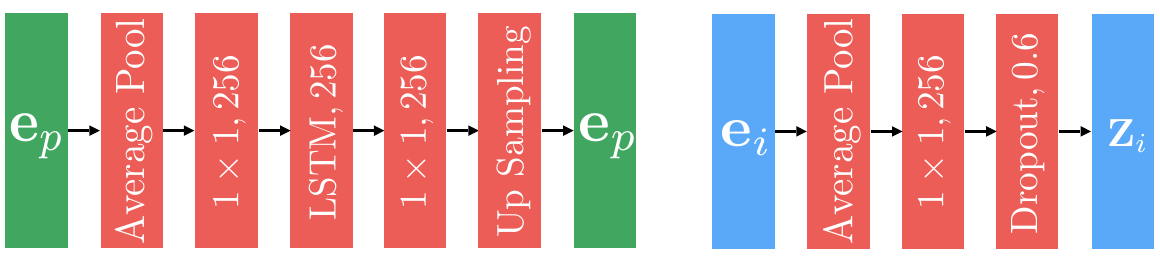}
\caption{{\bf Left:} the architecture of $f_{trn}$. We use average pooling to cut down the input dimension of LSTM and recover the dimension by upsampling. {\bf Right:} the architecture of $f_{cls}$. We set $\mathbf{z}_i \in \mathbb{R}^{256}$ to achieve a compact identity representation.}\label{fig:trn_cls}
\end{figure}

\subsection{The Design of $f_{cls}$}

The design of $f_cls$ is shown in Figure \ref{fig:trn_cls} (right). The purpose of $f_{cls}$ is to apply additional identity constraint on $\mathbf{e}_{i}$, so the identity and non-identity codes can be decoupled more completely. Specially, $f_{cls}$ takes $\mathbf{e}_{i}$ as  the input and output a $256d$ feature vector for the identity representation. Instead of using a very long feature vector in former face recognition networks \cite{TaigmanCVPR14}, {\em e.g.} $4096d$, we use a compact one, {\em e.g.} $256d$, to reduce the computational cost for efficient training \cite{SchroffCVPR15,SunCVPR15}. We apply $0.6$ dropout on the $256d$ vector to avoid overfitting. The vector is then followed by a fully-connected layer of $N$ neurons to output an one-hot vector for the identity prediction, where $N$ is the number of different subjects in training sequences. We use the cross-entropy loss defined in Equation \eqref{eq:clsloss} to train the identity task.

\begin{table*}[t]
\centering
\caption{The image and video datasets used in training and evaluation. We split AFLW and 300-VW into two parts for training and evaluation, respectively. LFW, Helen, LFPW, TF, and FM are used for training only. Note that LFW, TF, FM and 300-VW have both landmark and identity annotations; while the others have only landmark annotations.} \label{tab:dataset}
\begin{tabular}{c | c c c c | c  c c }
\toprule
& AFLW \cite{Koestinger11}  & LFW \cite{Gary14} & Helen \cite{LeECCV12} & LFPW \cite{BelhumeurCVPR11} & TF \cite{fgnet04} & FM \cite{PengICCV15} & 300-VW \cite{ShenICCVW15}  \\
\hline
in-the-wild setting & yes & yes & yes & yes & no & yes & yes \\
image number  & 21,080 & 12,007 & 2,330 & 1,035 & 500 & 2,150 & 114,000 \\
video number  & - & - & - & - & 5 & 6 & 114 \\ 
landmark annotation  & 21pt & 7pt & 194pt & 68pt & 68pt & 68pt & 68pt \\
subject number  & - & 5,371 & - & - & 1 & 6 & 105 \\
\hline
used in training & 16,864 & 12,007 & 2,330 & 1,035 & 0 & 0 & 90,000 \\
used in evaluation & 4,216 & 0 & 0 & 0 &  500 & 2150 & 24,000 \\
\bottomrule
\end{tabular}
\end{table*}

\section{Experiments}

We first introduce the datasets and settings. Then we carry out comprehensive module-wise study to validate the proposed method in various aspects. Finally, we compare our method with state-of-the-arts on both controlled and in-the-wild datasets.

\subsection{Datasets and Settings}

\textbf{Datasets.} We conduct our experiments on both image and video datasets. These datasets are widely used in face alignment and recognition. They present challenges in multiple aspects such as large pose, extensive expression, severe occlusion and dynamic illumination. Totally 7 datasets are used:
\begin{itemize}
\item Annotated Facial Landmarks in the Wild (AFLW)~\cite{Koestinger11}
\item Labeled Faces in the Wild (LFW)~\cite{Gary14}
\item Helen facial feature dataset (Helen)~\cite{LeECCV12,SagonasICCVW13}
\item Labeled Face Parts in the Wild (LFPW)~\cite{BelhumeurCVPR11,SagonasICCVW13}
\item Talking Face (TF)~\cite{fgnet04}
\item Face Movies (FM)~\cite{PengICCV15}
\item 300 face Videos in the Wild (300-VW)~\cite{ShenICCVW15}
\end{itemize}

We list configurations and setups of each dataset in Table \ref{tab:dataset}. Different datasets have different landmark annotation protocol. For Helen, LFPW, TF, FM and 300-VW, we follow \cite{SagonasICCVW13,Sagonas20163} to obtain both 68- and 7-landmark annotation. For AFLW, we generat 7-landmark annotations using the original 21 landmarks. The landmark annotation of LFW is given by \cite{Gary14}. For identity labels, we manually label all videos in TF, FM, and 300-VW. It is easy since the identity is unique is a given video.

AFLW and 300-VW have the largest number of labeled images. They are also more challenging than others due to the extensive variations. Therefore, we use them for both training and evaluation. More specifically, $80\%$ of the images in AFLW and $90$ out of $114$ videos in 300-VW are used for training, and the rest are used for evaluation. We sample videos to roughly cover the three different scenarios defined in \cite{ChrysosICCVW15}, \textit{i.e.} "Scenario 1", "Scenario 2" and "Scenario 3", corresponding to well-lit, mild unconstrained and completely unconstrained conditions.

We perform data augmentation by sampling ten variations from each image in the image training datasets. The sampling was achieved by random perturbation of scale ($0.9$ to $1.1$), rotation ($\pm 15^\circ$), translation ($7$ pixels), as well as horizontal flip. To generate sequential training data, we randomly sample 100 clips from each training video, where each clip has 10 frames. It is worth mentioning that no augmentation is applied on video training data to preserve the temporal consistency in the successive frames.

\textbf{Compared methods.} We compared the proposed method with both regression based and deep learning based approaches that reported state-of-the-art performance in unconstrained conditions. Totally 8 methods are compared:
\begin{itemize}
\item Discriminative Response Map Fitting (DRMF)~\cite{AsthanaCVPR13}
\item Explicit Shape Regression (ESR)~\cite{CaoIJCV14}
\item Supervised Descent Method (SDM)~\cite{XiongCVPR13}
\item Incremental Face Alignment (IFA)~\cite{AsthanaCVPR14}
\item Coarse-to-Fine Shape Searching (CFSS)~\cite{ZhuCVPR15}
\item Deep Convolutional Network Cascade (DCNC)~\cite{SunCVPR13}
\item Coarse-to-fine Auto-encoder Network (CFAN)~\cite{ZhangECCV14}
\item Deep Multi-task Learning (TCDCN)~\cite{ZhangTangECCV14}
\end{itemize}

For image-based evaluation, we follow \cite{AsthanaCVPR13} to provide a bounding box as the face detection output. For video-based evaluation, we follow \cite{PengICCV15} to utilize a tracking-by-detection protocol, where the face bounding box of the current frame is calculated according to the landmark of the previous frame.

\textbf{Training strategy.} Our approach is capable of end-to-end training. However, there are only 105 different subjects presented in 300-VW, which hardly provide sufficient supervision for the identity constraint. To make full use of all datasets, we conducted the training through three steps. {\bf First}, we pre-train the network without $f_{trn}$ and $f_{cls}$ using image-based datasets, {\it i.e.,} AFLW~\cite{Koestinger11}, Helen~\cite{LeECCV12} and LFPW~\cite{BelhumeurCVPR11}. {\bf Then}, $f_{cls}$ is engaged for identity constraint and fine-tuned together with other modules using image-based LFW~\cite{Gary14}. {\bf Finally}, $f_{trn}$ is triggered and the entire network is fine-tuned using video-based dataset, {\it i.e.} 300-VW~\cite{ShenICCVW15}.

\textbf{Experimental Settings.} In every frame, the initial response map $\mathbf{z}_0$ (Equation \eqref{eq:z}) is generated using the landmark prediction of the previous frame. Parameter $\lambda$ and $\gamma$ (Equation \eqref{eq:allloss}) are empirically set so the ratio of $\ell_{det} : \ell_{reg} : \ell_{cls}$ is roughly equal to $1:10:1$.

For training, we optimize the network parameters by using \textit{stochastic gradient descent} (SGD) with 0.9 momentum. We use fixed learning rate started at 0.01 and manually decreased it to an order of magnitude according to the validation accuracy. $f_{enc}$ is initialized using pre-trained weights of VGG-16 \cite{SimonyanCoRR14} or ResNet-152 \cite{he2016deep}. Other modules are initialized with Gaussian initialization \cite{JiaACMM14}. The training clips in a mini-batch have no identity overlap to avoid oscillations of the identity loss. We perform temporal recurrent learning in both forward and backward direction to double the usage of the sequential corpus.

For testing, we split 300-VW so that the training and testing videos do not have identity overlap (16 videos share 7 identities) to avoid overfitting. We use the inter-ocular distance to normalize the \textit{root mean square error} (RMSE) \cite{SagonasICCVW13} for accuracy evalutaion. A prediction with larger than $10\%$ mean error is reported as a failure \cite{ShenICCVW15}.

\subsection{Validation of Encoder-decoder Variants}
In Section \ref{sec:arch_encdec}, we proposed two different designs of encoder-decoder: {\bf (1)} VGGNet-based design with symmetrical encoder and decoder, which has been mainly investigated in our former conference paper \cite{peng2016recurrent}; and {\bf (2)} ResNet-based design with asymmetrical encoder, {\it i.e.}, the encoder is much deeper than the decoder. In particular, skip connections are incorporated in bridging the encoder and decoder with hierarchical spatial information at different resolutions.

We compared the performance of two encoder-decoder variants on AFLW~\cite{Koestinger11} and 300-VW~\cite{ShenICCVW15}. The results are reported in Table \ref{tab:encdec}. The results show that the ResNet-based design outperforms the VGGNet-based variant with a substantial margin in terms of fitting accuracy (mean error) and robustness (standard deviation). Much deeper layers, as well as the proposed skipping shortcuts, contribute a lot to the improvement. In addition, the ResNet-based encoder-decoder has very close computational cost to the VGGNet-based variant, {\it e.g.} the average fitting time per image/frame and the memory usage of a trained model, which should be attributed to the custom residual module design and the proposed asymmetrical encoder-decoder network.

\begin{table}[t]
\centering
\caption{Performance comparison of VGGNet-based and ResNet-based encoder-decoder Variants. Network configurations are described in Section \ref{sec:arch_encdec}. Row 1-2: image-based results on AFLW~\cite{Koestinger11}; Row 3-4: video-based results on 300-VW~\cite{ShenICCVW15}.} \label{tab:encdec}
\begin{tabular}{c | c c | c c}
\toprule
& Mean (\%) & Std (\%) & Time & Memory \\
\hline
VGGNet-based & 6.85 & 4.52  & 43.6$ms$ & 184$Mb$ \\ 
ResNet-based & 6.33 & 3.61  & 54.9$ms$ & 257$Mb$ \\
\hline 
\multicolumn{5}{c}{}\\[-0.5em] 
\hline
VGGNet-based & 5.16 & 2.57  & 42.5$ms$ & 184$Mb$ \\ 
ResNet-based & 4.75 & 2.10  & 56.2$ms$ & 257$Mb$ \\
\bottomrule
\end{tabular}
\end{table}

\begin{table}[t]
\centering
\caption{
Comparison of single-step detection or regression with the proposed recurrent detection-followed-by-regression on AFLW~\cite{Koestinger11}. The proposed method (Last Row) has the best performance especially in challenging settings.} \label{tab:srn_error}
\begin{tabular}{ c c c c c c }
\toprule
& \multicolumn{2}{c}{Common ($\%$)} & & \multicolumn{2}{c}{Challenging ($\%$)} \\
\cline{2-3} \cline{5-6}
& Error & Failure & & Error & Failure \\
\multicolumn{6}{c}{}\\[-0.95em]
\hline
Single-step Detection		& 6.05 & 4.62 &  & 8.14 & 12.4 \\
Single-step Regression		& 5.92 & 4.75 &  & 7.87 & 14.5 \\
Recurrent Det.+Det.			& 5.86 & 3.44 &  & 7.33 & 8.20 \\
Recurrent Det.+Reg.			& 5.71 & 3.30 &  & 6.97 & 8.75 \\
\bottomrule
\end{tabular}
\end{table}

\begin{table}[b]
\centering
\caption{Comparison of cascade and recurrent learning in the challenging settings of AFLW~\cite{Koestinger11}. The latter improves accuracy with a half memory usage of the former.} \label{tab:srn_error2}
\begin{tabular}{c | c c | c c}
\toprule
& Mean ($\%$) & Std ($\%$) & Memory \\
\hline
Cascade Det. \& Reg.  & 6.81 & 4.53  & 468$Mb$ \\ 
Recurrent Det. \& Reg. & 6.33 & 3.61  & 257$Mb$ \\
\bottomrule
\end{tabular}
\end{table}

\begin{table*}[t]
\centering
\caption{Validation of temporal recurrent learning on 300-VW \cite{SagonasICCVW13}. $f_{trn}$ helps to improve the tracking robustness (smaller std and lower failure rate), as well as the tracking accuracy (smaller mean error). The improvement is more significant in challenging settings of large pose and partial occlusion as demonstrated in Figure \ref{fig:trn_curve}.} \label{tab:trn_error}
\begin{tabular}{ c c c c c c c c c c c c }
\toprule
& \multicolumn{3}{c}{Common} & & \multicolumn{3}{c}{Challenging} & & \multicolumn{3}{c}{Full}\\
\cline{2-4} \cline{6-8} \cline{10-12}
& Mean (\%) & Std (\%) & Fail (\%) & & Mean (\%) & Std (\%) & Failure (\%) & & Mean (\%) & Std (\%) & Fail (\%)\\
\hline
w/o $f_{trn}$ 	& 4.52 & 2.24 & 3.48 &  & 6.27 & 5.33 & 13.3 &  & 5.83 & 3.42 & 6.43 \\
$f_{trn}$ 			& 4.21 & 1.85 & 1.71 &  & 5.64 & 3.28 & 5.40 &  & 5.25 & 2.15 & 2.82\\
\bottomrule
\end{tabular}
\end{table*}

\subsection{Validation of Spatial Recurrent Learning}
We validated the proposed spatial recurrent learning on the validation set of AFLW~\cite{Koestinger11}. To better investigate the benefits of spatial recurrent learning, we partitioned the validation set into two image groups according to the absolute value of the yaw angle: {\bf (1)} Common settings where $\text{yaw} \in [0^{\circ}$-$30^{\circ})$; and {\bf (2)} Challenging settings where $\text{yaw} \in (30^{\circ}$,$90^{\circ}]$. The training sets are ensembles of AFLW~\cite{Koestinger11}, Helen~\cite{LeECCV12} and LFPW~\cite{BelhumeurCVPR11} as described in Table \ref{tab:dataset}.

\bf Validation of detection-followed-by-regression.} To validate the proposed recurrent detection-followed-by-regression, we investigated four different network configurations:
\begin{itemize}
\item Single-step prediction using loss defined in Equation \eqref{eq:det_loss}; 
\item Single-step prediction using loss defined in Equation \eqref{eq:reg_loss};
\item Two-step recurrent detection-followed-by-detection; 
\item Two-step recurrent detection-followed-by-regression.
\end{itemize}

The mean fitting errors and failure rates are reported in Table \ref{tab:srn_error}. First, the results show that the two-step recurrent learning can instantly decrease the fitting error and failure rate, compared with either the single-step detection or regression. The improvement is more significant in challenging settings with large pose variations. Second, though landmark detection is more robust in challenging settings (low failure rate), it lacks the ability to predict precise locations (small fitting error) compared to landmark regression. This fact proves the effectiveness of the proposed recurrent detection-followed-by-regression.

{\bf Validation of recurrent learning.} We also conducted comparisons between the proposed spatial recurrent learning and the cascade models that are widely used in former approaches \cite{SunCVPR13,ZhangECCV14}. For a fair comparison, we implemented a two-step cascade variant to perform detection-followed-by-regression. Each network in the cascade has exactly the same architecture as the recurrent version. But there is no weight sharing among cascades. We fully trained the cascade networks using the same training set and validated the performance in challenging settings. 

The comparison is shown in Table \ref{tab:srn_error2}. Unsurprisingly, the spatial recurrent learning can improve the fitting accuracy. The underlying reason is taht the recurrent network learns the step-by-step fitting strategy jointly, while the cascade networks learn each step independently. It can better handle the challenging settings where the initial guess is usually far away from the ground truth. Moreover, the recurrent network with shared weights can instantly reduce the memory usage to one-half of the cascaded model.

\begin{figure}[t]
\centering
\includegraphics[width=0.45\textwidth]{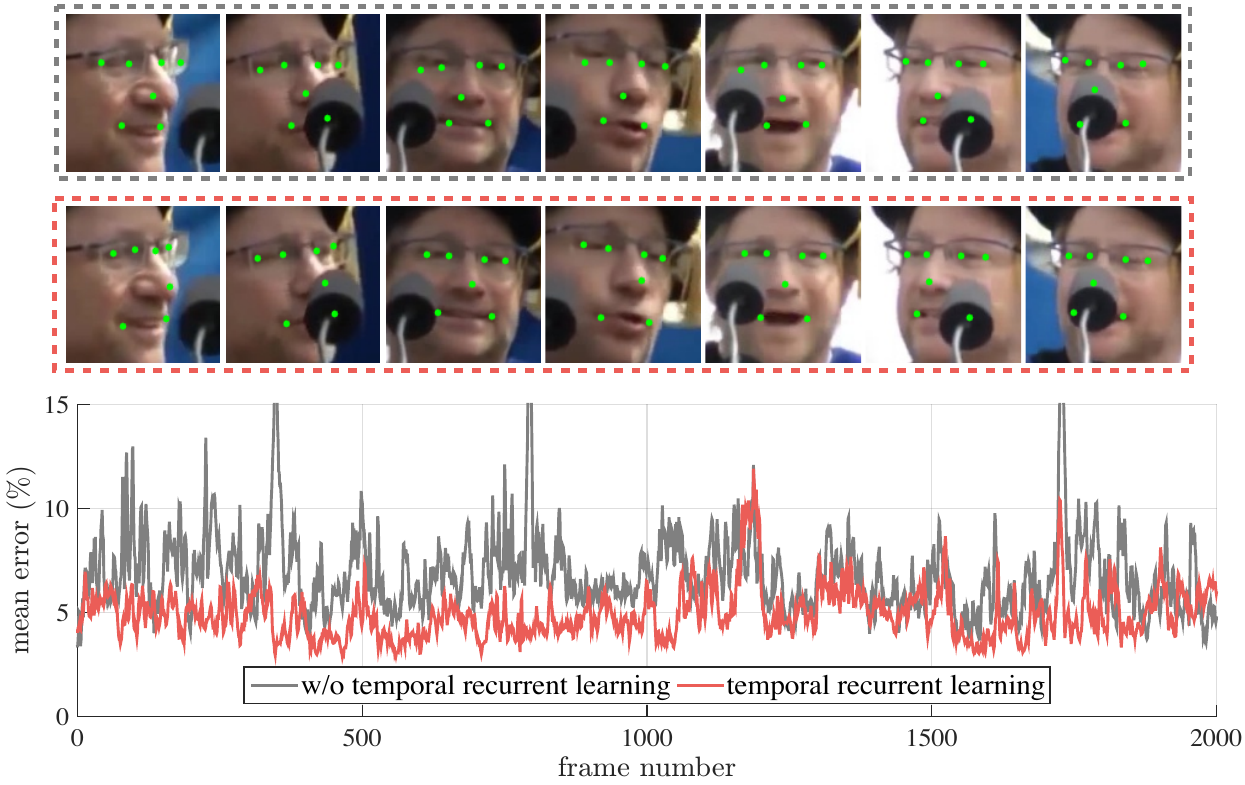}
\caption{Examples of temporal recurrent learning on 300-VW~\cite{SagonasICCVW13}. The tracked subject undergoes intensive pose and expression variations as well as severe partial occlusions. $f_{trn}$ substantially improves the tracking robustness (less variance) and fitting accuracy (low error), especially for landmarks on the nose tip and mouth corners.} \label{fig:trn_curve}
\end{figure}

\subsection{Validation of Temporal Recurrent Learning}
We validate the proposed temporal recurrent learning on the validation set of 300-VW~\cite{ShenICCVW15}. To better study the performance under different settings, we split the validation set into two groups: {\bf (1)} 9 videos in common settings that roughly match "Scenario 1"; and {\bf (2)} 15 videos in challenging settings that roughly match "Scenario 2" and "Scenario 3". The common, challenging and full sets were used for evaluation.

We implemented a variant of our approach that turns off the temporal recurrent learning $f_{trn}$. It was also pre-trained on the image training set and fine-tuned on the video training set. Since there was no temporal recurrent learning, we used frames instead of clips to conduct the fine-tuning which was performed for the same 50 epochs. We showed the result with and without temporal recurrent learning in Table \ref{tab:trn_error}.

For videos in common settings, the temporal recurrent learning achieves $6.8\%$ and $17.4\%$ improvement in terms of mean error and standard deviation respectively, while the failure rate is remarkably reduced by $50.8\%$. Temporal modeling produces better prediction by taking consideration of history observations. It may implicitly learn to model the motion dynamics in the hidden units from the training clips. 

For videos in challenging settings, the temporal recurrent learning won with even bigger margin. Without $f_{trn}$, it is hard to capture the drastic motion or changes in consecutive frames, which inevitably results in higher mean error, std and failure rate. Figure \ref{fig:trn_curve} shows an example where the subject exhibits intensive pose and expression variations as well as severe partial occlusions. The curve showed our recurrent model obviously reduced landmark errors, especially for landmarks on nose tip and mouth corners. The less oscillating error also suggests that $f_{trn}$ significantly improves the prediction stability over frames.

\begin{figure}[t]
\centering
\includegraphics[width=.897\linewidth]{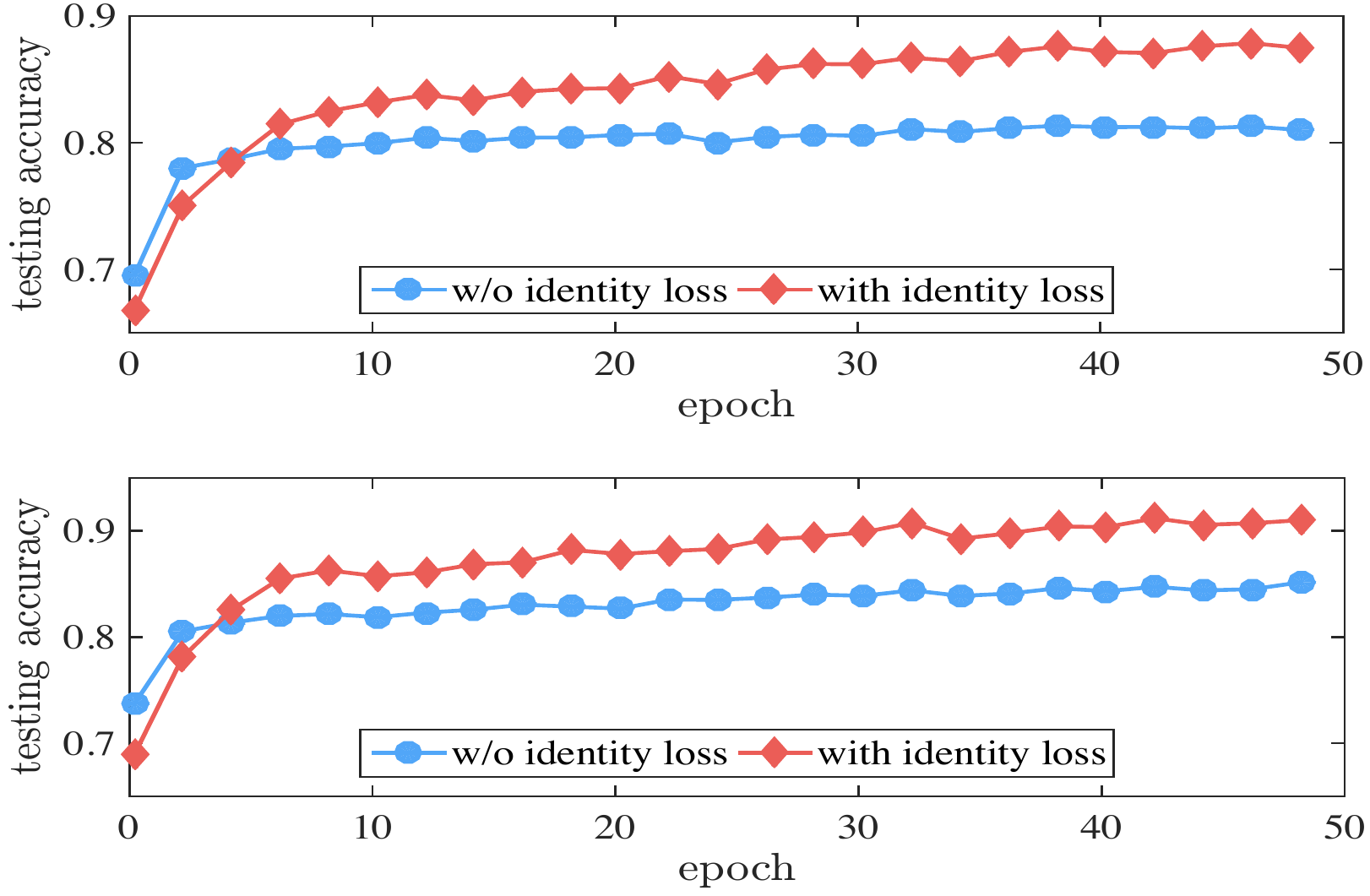}
\caption{Fitting accuracy of different facial components with respect to the number of training epochs on 300-VW~\cite{ShenICCVW15}. The proposed supervised identity disentangling helps to achieve a more complete factor decoupling in the bottleneck of the encoder-decoder, which yields better generalization capability and more accurate fitting results.}\label{fig:cls_curve}
\end{figure}

\begin{table*}[t]
\centering
\caption{Mean error comparison with state-of-the-arts on video-based validation sets: TF, FM, and 300-VW~\cite{SagonasICCVW13}. The top performance in each dataset is highlighted. Our approach achieves the best fitting accuracy on both controlled and unconstrained datasets.} \label{tab:soa_compare}.
\begin{tabular}{ c c c c c c c c c }
\toprule
& \multicolumn{3}{c}{7 landmarks} & & & \multicolumn{3}{c}{68 landmarks} \\
\cline{2-4} \cline{7-9} 
& TF~\cite{fgnet04} & FM~\cite{PengICCV15} & 300-VW~\cite{ShenICCVW15}  & & & TF~\cite{fgnet04} & FM~\cite{PengICCV15} & 300VW~\cite{ShenICCVW15} \\
\cline{1-4} \cline{6-9} 
DRMF~\cite{AsthanaCVPR13} & 4.43 & 8.53 & 9.16 & & ESR~\cite{CaoIJCV14}  & 3.49 & 6.74 & 7.09\\
ESR~\cite{CaoIJCV14}  & 3.81 & 7.58 & 7.83 & & SDM~\cite{XiongCVPR13}  & 3.80 & 7.38 & 7.25\\
SDM~\cite{XiongCVPR13} & 4.01 & 7.49 & 7.65 & & CFAN~\cite{ZhangECCV14}  & 3.31 &6.47 & 6.64\\
IFA~\cite{AsthanaCVPR14}  & 3.45 & 6.39 & 6.78 & & TCDCN~\cite{ZhangTangECCV14}  & 3.45 & 6.92 & 7.59 \\
DCNC~\cite{SunCVPR13} & 3.67 & 6.16 & 6.43 & & CFSS~\cite{ZhuCVPR15}  & 3.04 & 5.67 & 6.13\\
\cline{1-4} \cline{6-9} 
RED-Net (Ours) & $\bf{2.89}$ & $\bf{5.14}$ & $\bf{5.29}$ & & RED-Net (Ours) & $\bf{2.77}$ & $\bf{4.93}$ & $\bf{5.15}$\\ 
\bottomrule
\end{tabular}
\end{table*}

\subsection{Benefits of Supervised Identity Disentangling}

The supervised identity disentangling is proposed to better decouple the temporal-invariant and temporal-variant factors in the bottleneck of the encoder-decoder. This facilitates the temporal recurrent training, yielding better generalization and more accurate fittings at test time.

To study the effectiveness of the identity constraint, we removed $f_{cls}$ and follow the exact training steps. The testing accuracy comparison on the 300-VW~\cite{SagonasICCVW13} is shown in Figure \ref{fig:cls_curve}. The accuracy was calculated as the ratio of pixels that were correctly classified in the corresponding channel(s) of the response map.

The validation results of different facial components show similar trends: \textbf{(1)} The network demonstrates better generalization capability by using additional identity cues, which results in a more efficient training. For instance, after only 10 training epochs, the validation accuracy for landmarks located at the left eye reaches 0.84 with identity loss compared to 0.8 without identity loss. \textbf{(2)} The supervised identity information can substantially boost the testing accuracy. There is an approximately $9\%$ improvement by using the additional identity loss. It worth mentioning that, at the very beginning of the training (< 5 epochs), the network has inferior testing accuracy with supervised identity disentangling. It is because the suddenly added identity loss perturbs the backpropagation process. However, the testing accuracy with identity loss increases rapidly and outperforms the one without identity loss after only a few more training epochs.

\subsection{General Comparison with the State of the art}
We compared our framework with both traditional approaches and deep learning based approaches. The methods with hand-crafted features include: \textbf{(1)} DRMF \cite{AsthanaCVPR13}, \textbf{(2)} ESR \cite{CaoIJCV14}, \textbf{(3)} SDM \cite{XiongCVPR13}, \textbf{(4)} IFA \cite{AsthanaCVPR14}, and \textbf{(5)} PIEFA \cite{PengICCV15}. The deep learning based methods include: \textbf{(1)} DCNC  \cite{SunCVPR13}, \textbf{(2)} CFAN \cite{ZhangECCV14}, and \textbf{(3)} TCDCN \cite{ZhangTangECCV14}. All these methods were recently proposed and reported state-of-the-art performance. For fair comparison, we evaluated these methods in a tracking protocol: fitting result of current frame was used as the initial shape (DRMF, SDM and IFA) or the bounding box (ESR and PIEFA) in the next frame. The comparison was performed on both controlled, \textit{e.g.} Talking Face (TF) \cite{fgnet04}, and in-the-wild datasets, \textit{e.g.} Face Movie (FM) \cite{PengICCV15} and 300-VW \cite{ShenICCVW15}.

\begin{figure*}[t]
\centering
\subfloat{\includegraphics[width=1\linewidth]{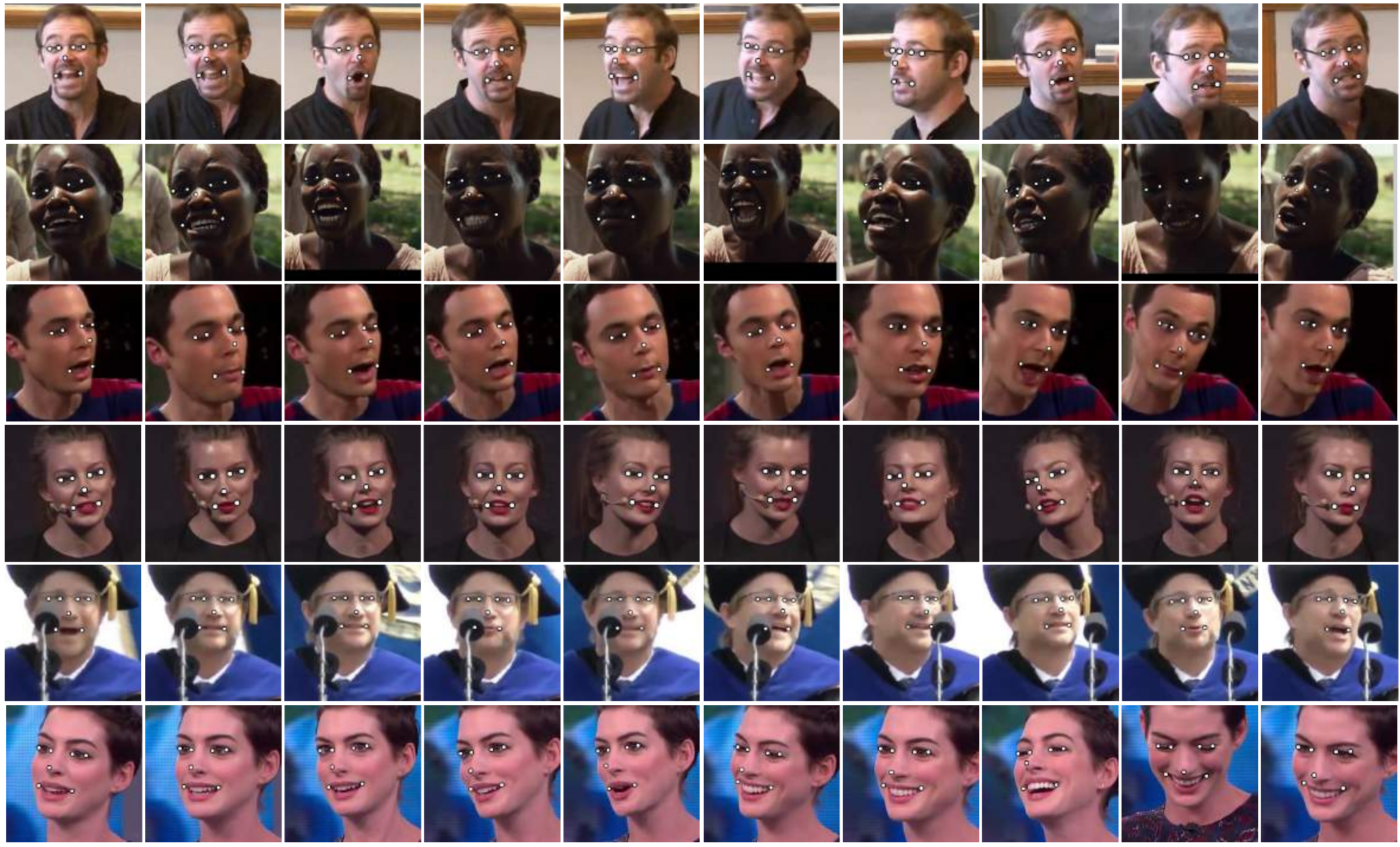}}\\
 \subfloat{\includegraphics[width=1\linewidth]{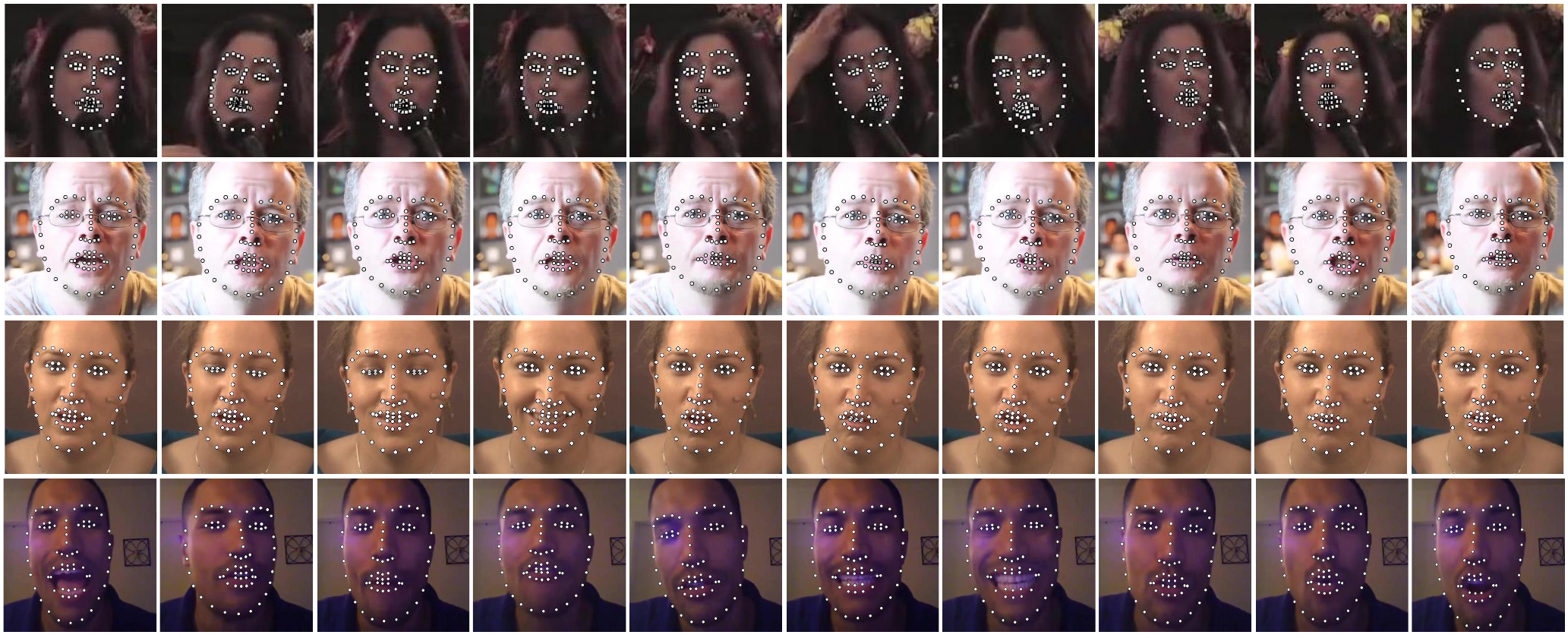}}
  \caption{Examples of 7-landmark ({\bf Row 1-6}) and 68-landmark ({\bf Row 7-10}) fitting results on FM~\cite{PengICCV15} and 300-VW~\cite{ShenICCVW15}. The proposed approach achieves robust and accurate fittings when the tracked subjects suffer from large pose/expression changes ({\bf Row 1, 3, 4, 6, 10}), illumination variations ({\bf Row 2, 8}) and partial occlusions ({\bf Row 5, 7}). }\label{fig:example}
\end{figure*}

We report the evaluation results for both 7 and 68 landmark setups in Table ~\ref{tab:soa_compare}. Our approach achieves state-of-the-art performance under both settings. It outperforms others with a substantial margin on all datasets under both 7-landmark and 68-landmark protocols. The performance gain is more significant on the challenging datasets (FM and 300-VW) than controlled dataset (TF). Our alignment model runs fairly fast, it takes around 40ms to process an image using a Tesla K40 GPU accelerator. Please refer to Figure \ref{fig:example} for fitting results of our approach on FM~\cite{PengICCV15} and 300-VW~\cite{ShenICCVW15}, which demonstrate the robust and accurate performance in wild conditions.

\section{Conclusion}
In this paper, we proposed a novel recurrent encoder-decoder network for real-time sequential face alignment. It utilizes spatial recurrency to train an end-to-end optimized coarse to fine landmark detection model. It decouples temporal-invariant and temporal-variant factors in the bottleneck of the network, and exploits recurrent learning at both spatial and temporal dimensions. Extensive experiments demonstrated the effectiveness of our framework and its superior performance. The proposed method provides a general framework that can be further applied to other localization-sensitive tasks, such as human pose estimation, object detection, scene classification, and others. 


\bibliographystyle{spmpsci}
\bibliography{manuscript}

\end{document}